\documentclass[onecolumn]{article}

\usepackage{graphicx}
\usepackage{amsmath} 
\usepackage{amssymb}  
\usepackage{lipsum}
\usepackage{multicol, multirow}
\usepackage{xcolor}
\usepackage{subfiles}
\usepackage[normalem]{ulem}

\newcommand{\nuc}{\newcommand}

\nuc{\ISd}{Itakura-Saito divergence}
\nuc{\pp}{point process}
\nuc{\Pp}{Point process}
\nuc{\Bb}{BB}
\nuc{\cvd}{covariance density}
\nuc{\psd}{power spectral density}
\nuc{\BD}{Bregman divergence}
\nuc{\ifunc}{intensity function}
\nuc{\mgf}{moment generating function}

\nuc{\Psp}{Poisson process}
\nuc{\rnp}{renewal process}
\nuc{\rp}{renewal process}
\nuc{\Rnp}{Renewal process}
\nuc{\Rp}{Renewal process}
\nuc{\Hp}{Hawkes process}
\nuc{\HL}{Hawkes-Laguerre}
\nuc{\Lt}{Laplace transform}
\nuc{\iLt}{inverse Laplace transform}
\nuc{\iets}{interevent times}
\nuc{\Gd}{Gamma distributed}
\nuc{\lhf}{likelihood function}
\nuc{\lh}{likelihood}
\nuc{\tiv}{time-invariant}
\nuc{\tv}{time-varying}
\nuc{\Bhatt}{Bhattacharyya}
\nuc{\Bnl}{Bernoulli}
\nuc{\cpi}{counting process increment}

\nuc{\cme}{concentrated matrix exponential}
\nuc{\se}{self-exciting}
\nuc{\AWf}{Abate-Whitt framework}

\nuc{\ul}{\underline}

\nuc{\Ra}{\Rightarrow}
\nuc{\La}{\Leftarrow}
\nuc{\LRa}{\Leftrightarrow}
\nuc{\toi}{\to\infty}
\nuc{\trid}{\triangledown}
\nuc{\triu}{\triangleup}

\nuc{\swn}{\ssum1^n}
\nuc{\swN}{\ssum1^N}
\nuc{\swP}{\ssum1^P}
\nuc{\swp}{\ssum1^p}
\nuc{\soN}{\ssum0^N}
\nuc{\swm}{\ssum1^m}
\nuc{\swM}{\ssum1^M}
\nuc{\siwn}{\ssum{i=1}^n}
\nuc{\siwnl}{\ssum{i=1}^{n_l}}
\nuc{\srwn}{\ssum{r=1}^n}
\nuc{\srwnl}{\ssum{r=1}^{n_l}}
\nuc{\srwnm}{\ssum{r=1}^{n_m}}
\nuc{\sawM}{\ssum{a=1}^M}
\nuc{\sbwM}{\ssum{b=1}^M}
\nuc{\siwm}{\ssum{i=1}^m}
\nuc{\skwm}{\ssum{k=1}^m}
\nuc{\siwM}{\ssum{i=1}^M}
\nuc{\skwM}{\ssum{k=1}^M}
\nuc{\sjwm}{\ssum{j=1}^m}
\nuc{\sjwM}{\ssum{j=1}^M}
\nuc{\slwM}{\ssum{l=1}^M}
\nuc{\smwM}{\ssum{m=1}^M}
\nuc{\sjwp}{\ssum{j=1}^p}
\nuc{\sjwP}{\ssum{j=1}^P}
\nuc{\siwK}{\ssum{i=1}^K}
\nuc{\swk}{\ssum1^k}
\nuc{\soi}{\ssum0^\infty}
\nuc{\swi}{\ssum1^\infty}
\nuc{\pwn}{\Pi_1^n}
\nuc{\pwk}{\Pi_1^k}
\nuc{\pwm}{\Pi_1^m}
\nuc{\prwn}{\Pi_{r=1}^n}

\nuc{\Stiltr}{\sum_{\ti<\tr}}
\nuc{\Sticlltrcm}{\sum_{\ticl<\trcm}}

\nuc{\Swn}{\sum_1^n}
\nuc{\Swm}{\sum_1^m}
\nuc{\SwM}{\sum_1^M}
\nuc{\Swk}{\sum_1^k}
\nuc{\Pwn}{\prod_1^n}
\nuc{\Pwm}{\prod_1^m}
\nuc{\Swp}{\sum_1^p}
\nuc{\SwP}{\sum_1^P}

\nuc{\intpi}{\int_{-\pi}^{\pi}}
\nuc{\intinfty}{\int_{-\infty}^{\infty}}
\nuc{\intoT}{\int_0^T}
\nuc{\intox}{\int_0^x}
\nuc{\intxi}{\int_x^\infty}
\nuc{\intot}{\int_0^t}
\nuc{\intoinfty}{\int_0^\infty}
\nuc{\intoi}{\int_0^\infty}

\nuc{\limto}{\lim_{t\to0}}
\nuc{\limTo}{\lim_{T\to0}}
\nuc{\limxo}{\lim_{x\to0}}
\nuc{\limso}{\lim_{s\to0}}
\nuc{\stoo}{s\to0}
\nuc{\stoi}{s\to\infty}
\nuc{\xtoi}{x\to\infty}
\nuc{\xtoo}{x\to0}
\nuc{\Ttoi}{T\to\infty}

\nuc{\limti}{\lim_{t\to\infty}}
\nuc{\limTi}{\lim_{T\to\infty}}
\nuc{\limxi}{\lim_{x\to\infty}}
\nuc{\limsi}{\lim_{s\to\infty}}

\nuc{\hlamt}{\hat{\lambda}(t)}

\nuc{\nid}{n_i^\delta}
\nuc{\Nid}{N_i^\delta}

\nuc{\intsum}{\ssum0^\infty\int_{R_n(T)}}
\nuc{\ints}{\ssum0^\infty\int_{R_n(T)}}

\nuc{\wos}{\frac1s}
\nuc{\wox}{\frac1x}
\nuc{\wot}{\frac1t}
\nuc{\woT}{\frac1T}
\nuc{\wo}[1]{\frac1{#1}}

\nuc{\cH}{\mathcal{H}}
\nuc{\cL}{\mathcal{L}}
\nuc{\cK}{\mathcal{K}}
\nuc{\cl}{\mathcal{l}}
\nuc{\cI}{\mathcal{I}}
\nuc{\bbR}{\mathbb{R}}
\nuc{\bbN}{\mathbb{N}}
\nuc{\bx}{{\bf x}}
\nuc{\by}{{\bf y}}

\nuc{\mysection}[1]{\section{#1}\setcounter{equation}{0}}
\nuc{\mypro}[2]{\noindent{\it Proof of #1. }{#2}\hfill$\square$}

\nuc{\eq}[1]{\begin{align*}#1\end{align*}}
\nuc{\eqn}[1]{\begin{align}#1\end{align}}
\nuc{\bmat}[1]{\begin{bmatrix}#1\end{bmatrix}}
\nuc{\mat}[1]{\begin{matrix}#1\end{matrix}}
\nuc{\smat}[1]{\begin{smallmatrix}#1\end{smallmatrix}}

\nuc{\theo}[1]{\begin{theorem}#1\end{theorem}}
\nuc{\lem}[1]{\begin{lemma}#1\end{lemma}}
\nuc{\defi}[1]{\begin{definition}#1\end{definition}}
\nuc{\exa}[1]{\begin{example}#1\end{example}}
\nuc{\cor}[1]{\begin{corollary}#1\end{corollary}}
\nuc{\prop}[1]{\begin{proposition}#1\end{proposition}}
\nuc{\res}[1]{\begin{result}#1\end{result}}
\nuc{\pro}[1]{\begin{proof}#1\end{proof}}
\nuc{\cas}[1]{\begin{cases}#1\end{cases}}
\nuc{\arr}[2]{\begin{array}{#1}#2\end{array}}
\nuc{\bra}[1]{\left(#1\right)}
\nuc{\sqbra}[1]{\left[#1\right]}
\nuc{\ang}[1]{\langle#1\rangle}
\nuc{\Ver}[1]{\lVert#1\rVert}
\nuc{\ver}[1]{\lvert#1\rvert}
\nuc{\ssum}[1]{{\textstyle\sum}_{#1}}
\nuc{\sprod}[1]{{\textstyle\prod}_{#1}}

\nuc{\sbij}[1]{#1_{ij}}
\nuc{\sbbij}[1]{\bar #1_{ij}}
\nuc{\sbi}[1]{#1_{i}}
\nuc{\sbbi}[1]{\bar #1_{i}}
\nuc{\sbj}[1]{#1_{j}}
\nuc{\sbbj}[1]{\bar #1_{j}}
\nuc{\nfd}[1]{#1^{(n)}}
\nuc{\limo}[1]{\lim_{#1\to0}}
\nuc{\limi}[1]{\lim_{#1\to\infty}}

\nuc{\deri}[2]{\frac{d#1}{d#2}}
\nuc{\ddx}{\frac d{dx}}
\nuc{\ddt}{\frac d{dt}}
\nuc{\ddtau}{\frac d{d\tau}}

\nuc{\ppx}{\frac \partial{ \partialx}}
\nuc{\ppt}{\frac  \partial{ \partialt}}
\nuc{\pptau}{\frac \partial{ \partial\tau}}

\nuc{\grad}{\bigtriangledown}

\newcommand{\nlist}[1]{\begin{enumerate}#1\end{enumerate}}
\newcommand{\blist}[1]{\begin{itemize}#1\end{itemize}}
\nuc{\itum}[1]{\item[(#1)]}
\nuc{\ita}{\itum{a}}
\nuc{\itb}{\itum{b}}
\nuc{\itc}{\itum{c}}
\nuc{\itd}{\itum{d}}
\nuc{\ite}{\itum{e}}
\nuc{\itf}{\itum{f}}
\nuc{\itg}{\itum{g}}
\nuc{\ith}{\itum{h}}

\nuc{\ai}{a_i}
\nuc{\ak}{a_k}
\nuc{\aj}{a_j}
\nuc{\aij}{a_{ij}}
\nuc{\akl}{a_{kl}}
\nuc{\akj}{a_{kj}}
\nuc{\abi}{\bar a_i}
\nuc{\abk}{\bar a_k}
\nuc{\abj}{\bar a_j}
\nuc{\abij}{\bar a_{ij}}
\nuc{\abkj}{\bar a_{kj}}
\nuc{\alpi}{\alpha_{i}}
\nuc{\alpk}{\alpha_{k}}
\nuc{\alpj}{\alpha_{j}}
\nuc{\alpij}{\alpha_{ij}}
\nuc{\alpkj}{\alpha_{kj}}
\nuc{\alpab}{\alpha_{ab}}
\nuc{\Ahw}{\hat A_1}
\nuc{\Aho}{\hat A_0}
\nuc{\ajcl}{a_{j,l}}
\nuc{\ajclcm}{a_{j,l,m}}

\nuc{\Bi}{B_i}
\nuc{\Bj}{B_j}
\nuc{\Bij}{B_{ij}}
\nuc{\Bkj}{B_{kj}}
\nuc{\bij}{b_{ij}}
\nuc{\bkj}{b_{kj}}
\nuc{\Bbi}{\bar B_i}
\nuc{\Bbj}{\bar B_j}
\nuc{\Bbij}{\bar B_{ij}}
\nuc{\Bbkj}{\bar B_{kj}}
\nuc{\beti}{\beta_i}
\nuc{\betj}{\beta_j}
\nuc{\betk}{\beta_k}
\nuc{\betaa}{\beta_a}
\nuc{\betb}{\beta_b}
\nuc{\betij}{\beta_{ij}}
\nuc{\betkl}{\beta_{kl}}
\nuc{\betkj}{\beta_{kj}}
\nuc{\betab}{\beta_{ab}}
\nuc{\betjcl}{\beta_{j,l}}
\nuc{\betjclcm}{\beta_{j,l,m}}
\nuc{\Bhw}{\hat B_1}
\nuc{\Bho}{\hat B_0}

\nuc{\ci}{c_i}
\nuc{\cj}{c_j}
\nuc{\ck}{c_k}
\nuc{\ca}{c_a}
\nuc{\cb}{c_b}

\nuc{\cm}{c_m}
\nuc{\cij}{c_{ij}}
\nuc{\ckj}{c_{kj}}
\nuc{\cab}{c_{ab}}
\nuc{\cbi}{\bar c_i}
\nuc{\cbj}{\bar c_j}
\nuc{\Ci}{C_i}
\nuc{\Cj}{C_j}
\nuc{\Cij}{C_{ij}}
\nuc{\Cbi}{\bar C_i}
\nuc{\Cbj}{\bar C_j}
\nuc{\Chw}{\hat C_1}
\nuc{\Cho}{\hat C_0}

\nuc{\Di}{D_i}
\nuc{\Dj}{D_j}
\nuc{\Dbi}{\bar D_i}
\nuc{\Dbj}{\bar D_j}
\nuc{\Dij}{D_{ij}}
\nuc{\Dkj}{D_{kj}}
\nuc{\Dab}{D_{ab}}
\nuc{\dij}{d_{ij}}
\nuc{\dab}{d_{ab}}
\nuc{\dkj}{d_{kj}}
\nuc{\Dbij}{\bar D_{ij}}
\nuc{\Dbkj}{\bar D_{kj}}
\nuc{\Dbab}{\bar D_{ab}}

\nuc{\et}{e_t}

\nuc{\fk}{f_k}
\nuc{\Fi}{F_i}
\nuc{\Fk}{F_k}
\nuc{\Fj}{F_j}
\nuc{\Fa}{F_a}
\nuc{\Fb}{F_b}
\nuc{\Fm}{F_m}
\nuc{\Ft}{F_t}
\nuc{\Fbi}{\bar F_i}
\nuc{\Fbj}{\bar F_j}
\nuc{\Fba}{\bar F_a}
\nuc{\Fbb}{\bar F_b}
\nuc{\Fij}{F_{ij}}
\nuc{\Fkj}{F_{kj}}
\nuc{\Fab}{F_{ab}}
\nuc{\fij}{f_{ij}}
\nuc{\fkj}{f_{kj}}
\nuc{\fab}{f_{ab}}
\nuc{\Fbij}{\bar F_{ij}}
\nuc{\Fbkj}{\bar F_{kj}}
\nuc{\Fbab}{\bar F_{ab}}
\nuc{\Fwtu}{F_{12}}
\nuc{\Fbwtu}{\bar F_{12}}
\nuc{\Fpct}{F_{p,t}}

\nuc{\gam}{\gamma}
\nuc{\Gam}{\Gamma}
\nuc{\gami}{\gam_i}
\nuc{\gamk}{\gam_k}
\nuc{\gamj}{\gam_j}
\nuc{\gamij}{\gam_{ij}}
\nuc{\gamkj}{\gam_{kj}}
\nuc{\gamab}{\gam_{ab}}
\nuc{\Gi}{G_i}
\nuc{\Gk}{G_k}
\nuc{\Gj}{G_j}
\nuc{\Ga}{G_a}
\nuc{\Gb}{G_b}
\nuc{\Gbi}{\bar G_i}
\nuc{\Gbj}{\bar G_j}
\nuc{\Gij}{G_{ij}}
\nuc{\Gkj}{G_{kj}}
\nuc{\Gab}{G_{ab}}
\nuc{\Gbab}{\bar G_{ab}}
\nuc{\Gbij}{\bar G_{ij}}
\nuc{\Gbkj}{\bar G_{kj}}
\nuc{\Gwtu}{G_{12}}
\nuc{\Gbwtu}{\bar G_{12}}

\nuc{\hi}{h_i}
\nuc{\hk}{h_k}
\nuc{\hj}{h_j}
\nuc{\ha}{h_a}
\nuc{\hb}{h_b}
\nuc{\hd}{h_d}
\nuc{\hbi}{\bar h_i}
\nuc{\hbk}{\bar h_k}
\nuc{\hbj}{\bar h_j}
\nuc{\hba}{\bar h_a}
\nuc{\hbb}{\bar h_b}
\nuc{\hij}{h_{ij}}
\nuc{\hbij}{\bar h_{ij}}
\nuc{\hkj}{h_{kj}}
\nuc{\hbkj}{\bar h_{kj}}
\nuc{\hab}{h_{ab}}
\nuc{\hbab}{\bar h_{ab}}
\nuc{\hjcm}{h_{j,m}}
\nuc{\hjcl}{h_{j,l}}
\nuc{\hjclcm}{h_{j,l,m}}
\nuc{\Hi}{H_i}
\nuc{\Hk}{H_k}
\nuc{\Hj}{H_j}
\nuc{\Hbi}{\bar H_i}
\nuc{\Hbk}{\bar H_k}
\nuc{\Hbj}{\bar H_j}
\nuc{\Hij}{H_{ij}}
\nuc{\Hbij}{\bar H_{ij}}
\nuc{\Hkj}{H_{kj}}
\nuc{\Hbkj}{\bar H_{kj}}
\nuc{\Hab}{H_{kj}}
\nuc{\Hbab}{\bar H_{ab}}
\nuc{\Hwtu}{H_{12}}
\nuc{\Hbwtu}{\bar H_{12}}
\nuc{\Hinf}{\cH_\infty}
\nuc{\cHTicj}{\cH_{T,i,j}}

\nuc{\kapij}{\kappa_{ij}}
\nuc{\kapbij}{\bar\kappa_{ij}}
\nuc{\kapkj}{\kappa_{kj}}
\nuc{\kapbkj}{\bar\kappa_{kj}}
\nuc{\kapab}{\kappa_{ab}}
\nuc{\kapbab}{\bar\kappa_{ab}}
\nuc{\kapji}{\kappa_{ji}}
\nuc{\kapbji}{\bar\kappa_{ji}}
\nuc{\Ki}{K_i}
\nuc{\Kk}{K_k}
\nuc{\Kj}{K_j}
\nuc{\Kt}{K_t}
\nuc{\Kbi}{\bar K_i}
\nuc{\Kbk}{\bar K_k}
\nuc{\Kbj}{\bar K_j}
\nuc{\Kij}{K_{ij}}
\nuc{\Kbij}{\bar K_{ij}}
\nuc{\Kkj}{K_{kj}}
\nuc{\Kbkj}{\bar K_{kj}}
\nuc{\Kab}{K_{ab}}
\nuc{\Kbab}{\bar K_{ab}}
\nuc{\Kji}{K_{ji}}
\nuc{\Kbji}{\bar K_{ji}}
\nuc{\Kwtu}{K_{12}}
\nuc{\Kbwtu}{\bar K_{12}}

\nuc{\lam}{\lambda}
\nuc{\Lam}{\Lambda}
\nuc{\lami}{\lambda_{i}}
\nuc{\lamk}{\lambda_{k}}
\nuc{\lamj}{\lambda_{j}}
\nuc{\lama}{\lambda_{a}}
\nuc{\lamb}{\lambda_{b}}
\nuc{\lamm}{\lambda_{m}}
\nuc{\laml}{\lambda_{l}}
\nuc{\lamij}{\lambda_{ij}}
\nuc{\lamkj}{\lambda_{kj}}
\nuc{\lamab}{\lambda_{ab}}
\nuc{\lamicj}{\lambda_{i,j}}
\nuc{\lambij}{\bar\lambda_{ij}}
\nuc{\lambab}{\bar\lambda_{ab}}
\nuc{\Lami}{\Lambda_{i}}
\nuc{\Lamk}{\Lambda_{k}}
\nuc{\Lama}{\Lambda_{a}}
\nuc{\Lamj}{\Lambda_{j}}
\nuc{\Lamij}{\Lambda_{ij}}
\nuc{\Lamkj}{\Lambda_{kj}}
\nuc{\Lamab}{\Lambda_{ab}}
\nuc{\Lamicj}{\Lambda_{i,j}}
\nuc{\Lambij}{\bar\Lambda_{ij}}

\nuc{\lamTt}{\lam_t^T}
\nuc{\lamtgT}{\lam_{t|T}}
\nuc{\lamTu}{\lam_u^T}
\nuc{\lamugT}{\lam_{u|T}}
\nuc{\lamicjclcm}{\lam_{i,j,l,m}}
\nuc{\lamTtcu}{\lam_{t,u}^T}
\nuc{\lamtcugT}{\lam_{t,u|T}}
\nuc{\lamTtpw}{\lam^T_{t+1}}
\nuc{\lamtpwgT}{\lam_{t+1|T}}
\nuc{\LamTt}{\Lam_t^T}
\nuc{\LamtgT}{\Lam_{t|T}}
\nuc{\LamTu}{\Lam_u^T}
\nuc{\LamugT}{\Lam_{u|T}}
\nuc{\LamTtpw}{\Lam^T_{t+1}}
\nuc{\LamtpwgT}{\Lam_{t+1|T}}
\nuc{\LamTtcu}{\Lam_{t,u}^T}
\nuc{\LamtcugT}{\Lam_{t,u|T}}

\nuc{\Li}{L_i}
\nuc{\Lk}{L_k}
\nuc{\Lj}{L_j}
\nuc{\cLa}{\cL_a}
\nuc{\cLb}{\cL_b}

\nuc{\Lb}{L_b}
\nuc{\Lbi}{\bar L_i}
\nuc{\Lbk}{\bar L_k}
\nuc{\Lbj}{\bar L_j}
\nuc{\Lij}{L_{ij}}
\nuc{\Lkj}{L_{kj}}
\nuc{\Lab}{L_{ab}}
\nuc{\Licj}{L_{i,j}}
\nuc{\Lbij}{\bar L_{ij}}
\nuc{\Lbkj}{\bar L_{kj}}
\nuc{\Lwtu}{L_{12}}
\nuc{\Lbwtu}{\bar L_{12}}

\nuc{\mi}{m_i}
\nuc{\mk}{m_k}
\nuc{\mj}{m_j}
\nuc{\mij}{m_{ij}}
\nuc{\mkj}{m_{kj}}
\nuc{\mab}{m_{ab}}
\nuc{\Mij}{M_{ij}}
\nuc{\Mkj}{M_{kj}}
\nuc{\Mab}{M_{ab}}
\nuc{\Mfij}{M_{\fij}}
\nuc{\mui}{\mu_i}
\nuc{\muk}{\mu_k}
\nuc{\muj}{\mu_j}
\nuc{\mua}{\mu_a}
\nuc{\mub}{\mu_b}
\nuc{\muij}{\mu_{ij}}
\nuc{\mukj}{\mu_{kj}}
\nuc{\muab}{\mu_{ab}}
\nuc{\mbi}{\bar m_i}
\nuc{\mbk}{\bar m_k}
\nuc{\mbj}{\bar m_j}
\nuc{\mbij}{\bar m_{ij}}
\nuc{\mbkj}{\bar m_{kj}}

\nuc{\NoT}{N_0^T}
\nuc{\nut}{\nu_t}
\nuc{\nl}{n_l}
\nuc{\nm}{n_m}
\nuc{\Nt}{N_t}
\nuc{\NT}{N_T}
\nuc{\nT}{n_T}
\nuc{\nicj}{n_{i,j}}
\nuc{\Nu}{N_u}
\nuc{\Nucm}{N_{u,m}}
\nuc{\Nucl}{N_{u,l}}
\nuc{\Ntcl}{N_{t,l}}
\nuc{\Not}{N_0^t}
\nuc{\Notm}{N_0^{t_-}}

\nuc{\ome}{\omega}
\nuc{\Ome}{\Omega}
\nuc{\omekl}{\ome_{kl}}
\nuc{\omeal}{\ome_{al}}
\nuc{\omet}{\ome_t}
\nuc{\Omeicj}{\Ome_{i,j}}
\nuc{\Omebicj}{\bar\Ome_{i,j}}

\nuc{\pbi}{\bar p_i}
\nuc{\pbk}{\bar p_k}
\nuc{\pbj}{\bar p_j}
\nuc{\pba}{\bar p_a}
\nuc{\pbb}{\bar p_b}
\nuc{\pj}{p_j}
\nuc{\pk}{p_k}
\nuc{\pa}{p_a}
\nuc{\pb}{p_b}
\nuc{\pij}{p_{ij}}
\nuc{\pkj}{p_{kj}}
\nuc{\pab}{p_{ab}}
\nuc{\pbar}{\bar p}
\nuc{\pbij}{\bar p_{ij}}
\nuc{\pbkj}{\bar p_{kj}}
\nuc{\pbab}{\bar p_{ab}}
\nuc{\ptij}{\tilde p_{ij}}
\nuc{\pwtu}{p_{12}}
\nuc{\pbwtu}{\bar p_{12}}

\nuc{\Picj}{P_{i,j}}
\nuc{\PTicj}{P^T_{i,j}}
\nuc{\PTicipw}{P^T_{i,i+w}}
\nuc{\PTicipk}{P^T_{i,i+k}}

\nuc{\Pt}{P_t}
\nuc{\Pu}{P_u}
\nuc{\Ptcu}{P_{t,u}}
\nuc{\Ptck}{P_{t,k}}
\nuc{\PTtcu}{P^T_{t,u}}
\nuc{\PtcugT}{P_{t,u|T}}
\nuc{\Pstcu}{P^s_{t,u}}
\nuc{\Ptcugs}{P_{t,u|s}}
\nuc{\PTtctpw}{P^T_{t,t+w}}
\nuc{\PTtctpk}{P^T_{t,t+k}}
\nuc{\Ptt}{P^t_t}
\nuc{\Ptgt}{P_{t|t}}
\nuc{\PtgT}{P_{t|T}}
\nuc{\Pst}{P^s_t}
\nuc{\Ptgs}{P_{t|s}}
\nuc{\PTt}{P^T_t}
\nuc{\Ptmwt}{P^{t-1}_t}
\nuc{\Ptgtmw}{P_{t|t-1}}
\nuc{\Po}{P_0}
\nuc{\Poi}{P_0^{-1}}

\nuc{\Qbi}{\bar Q_i}
\nuc{\Qbk}{\bar Q_k}
\nuc{\Qbj}{\bar Q_j}
\nuc{\Qj}{Q_j}
\nuc{\Qij}{Q_{ij}}
\nuc{\Qkj}{Q_{kj}}
\nuc{\Qab}{Q_{ab}}
\nuc{\Qm}{Q_m}
\nuc{\Qjclcm}{Q_{j,l,m}}
\nuc{\Qbij}{\bar Q_{ij}}
\nuc{\Qwtu}{Q_{12}}
\nuc{\Qbwtu}{\bar Q_{12}}
\nuc{\qbi}{\bar q_i}
\nuc{\qbk}{\bar q_k}
\nuc{\qbj}{\bar q_j}
\nuc{\qba}{\bar q_a}
\nuc{\qbb}{\bar q_b}
\nuc{\qj}{q_j}
\nuc{\qij}{q_{ij}}
\nuc{\qbij}{\bar q_{ij}}
\nuc{\qkj}{q_{kj}}
\nuc{\qbkj}{\bar q_{kj}}
\nuc{\qab}{q_{ab}}
\nuc{\qbab}{\bar q_{ab}}
\nuc{\qwtu}{q_{12}}
\nuc{\qbwtu}{\bar q_{12}}

\nuc{\Rbi}{\bar R_i}
\nuc{\Rbj}{\bar R_j}
\nuc{\Ri}{R_i}
\nuc{\Rj}{R_j}
\nuc{\Rt}{R_t}
\nuc{\Rij}{R_{ij}}
\nuc{\Rkj}{R_{kj}}
\nuc{\Rab}{R_{ab}}
\nuc{\Rjclcm}{R_{j,l,m}}
\nuc{\rij}{r_{ij}}
\nuc{\Rbij}{\bar R_{ij}}
\nuc{\Rbkj}{\bar R_{kj}}
\nuc{\Rbab}{\bar R_{ab}}
\nuc{\Rwtu}{R_{12}}
\nuc{\Rbwtu}{\bar R_{12}}
\nuc{\rhoi}{\rho_{i}}
\nuc{\rhok}{\rho_{k}}
\nuc{\rhoj}{\rho_{j}}
\nuc{\rhoa}{\rho_{a}}
\nuc{\rhob}{\rho_{b}}
\nuc{\rhoij}{\rho_{ij}}
\nuc{\rhokj}{\rho_{kj}}
\nuc{\rhoab}{\rho_{ab}}
\nuc{\Rect}{R_{e,t}}

\nuc{\Si}{S_i}
\nuc{\Sk}{S_k}
\nuc{\Sj}{S_j}
\nuc{\Sa}{S_a}
\nuc{\Sb}{S_b}
\nuc{\Sbar}{\bar S}
\nuc{\Sbi}{\bar S_i}
\nuc{\Sbk}{\bar S_k}
\nuc{\Sbj}{\bar S_j}
\nuc{\Sba}{\bar S_a}
\nuc{\Sbb}{\bar S_b}
\nuc{\Sij}{S_{ij}}
\nuc{\Sbij}{\bar S_{ij}}
\nuc{\Skj}{S_{kj}}
\nuc{\Sbkj}{\bar S_{kj}}
\nuc{\Sab}{S_{ab}}
\nuc{\Sbab}{\bar S_{ab}}
\nuc{\Swtu}{S_{12}}
\nuc{\Sww}{S_{11}}
\nuc{\Swo}{S_{10}}
\nuc{\Sow}{S_{01}}
\nuc{\Soo}{S_{00}}
\nuc{\Swwf}{S_{11}^f}
\nuc{\Swof}{S_{10}^f}
\nuc{\Sowf}{S_{01}^f}
\nuc{\Soof}{S_{00}^f}
\nuc{\Swwb}{S_{11}^b}
\nuc{\Swob}{S_{10}^b}
\nuc{\Sowb}{S_{01}^b}
\nuc{\Soob}{S_{00}^b}
\nuc{\Sbwtu}{\bar S_{12}}
\nuc{\Sigxx}{\Sigma{xx}}
\nuc{\Sigx}{\Sigma{x}}
\nuc{\Sigy}{\Sigma{y}}
\nuc{\Sigxy}{\Sigma{xy}}
\nuc{\Sigyx}{\Sigma{yx}}
\nuc{\Sigyy}{\Sigma{yy}}

\nuc{\Sx}{S_{x}}
\nuc{\Sy}{S_{x}}
\nuc{\Sxx}{S_{xx}}
\nuc{\Sxy}{S_{xy}}
\nuc{\Syx}{S_{yx}}
\nuc{\Syy}{S_{yy}}
\nuc{\Szz}{S_{zz}}
\nuc{\Szx}{S_{zx}}
\nuc{\Sxxf}{S_{xx}^f}
\nuc{\Sxyf}{S_{xy}^f}
\nuc{\Syxf}{S_{yx}^f}
\nuc{\Syyf}{S_{yy}^f}
\nuc{\Sxxb}{S_{xx}^b}
\nuc{\Sxyb}{S_{xy}^b}
\nuc{\Syxb}{S_{yx}^b}
\nuc{\Syyb}{S_{yy}^b}
\nuc{\Sjclcm}{S_{j,l,m}}
\nuc{\sigi}{\sigma_i}
\nuc{\sigk}{\sigma_k}
\nuc{\siga}{\sigma_a}
\nuc{\sigb}{\sigma_b}

\nuc{\tn}{t_n}
\nuc{\tr}{t_r}
\nuc{\ti}{t_i}
\nuc{\ticl}{t_{i,l}}
\nuc{\trcm}{t_{r,m}}
\nuc{\twnm}{t_1^{n_m}}
\nuc{\twnl}{t_1^{n_l}}
\nuc{\twclnl}{t_{1,l}^{n_l}}
\nuc{\trmw}{t_{r-1}}
\nuc{\trpw}{t_{r+1}}
\nuc{\thehw}{\hat\theta_1}
\nuc{\theho}{\hat\theta_0}
\nuc{\thek}{\theta_k}
\nuc{\thej}{\theta_j}
\nuc{\thea}{\theta_a}
\nuc{\theb}{\theta_b}
\nuc{\theab}{\theta_{ab}}
\nuc{\thekj}{\theta_{kj}}
\nuc{\dtri}{\tr - \ti}
\nuc{\dtrcmicl}{\trcm - \ticl}
\nuc{\dTtr}{T - \tr}
\nuc{\dTti}{T - \ti}
\nuc{\dTticl}{T - \ticl}
\nuc{\dtr}{\trpw-\tr}
\nuc{\dtrcm}{t_{r+1,m}-\trcm}
\nuc{\tauwnicj}{\tau_1^{\nicj}}
\nuc{\tauicj}{\tau_{i,j}}
\nuc{\tauicjcl}{\tau_{i,j,l}}
\nuc{\twn}{t_1^{n}}
\nuc{\Twn}{T_1^n}
\nuc{\TwN}{T_1^{N_T}}
\nuc{\TwNT}{T_1^{N_T}}

\nuc{\ui}{u_i}
\nuc{\uk}{u_k}
\nuc{\uj}{u_j}
\nuc{\uij}{u_{ij}}
\nuc{\ukj}{u_{kj}}
\nuc{\uab}{u_{ab}}
\nuc{\ubi}{\bar u_i}
\nuc{\ubj}{\bar u_j}
\nuc{\Ui}{U_i}
\nuc{\Uj}{U_j}
\nuc{\Uij}{U_{ij}}
\nuc{\Ubi}{\bar U_i}
\nuc{\Ubj}{\bar U_j}

\nuc{\vi}{v_i}
\nuc{\vj}{v_j}
\nuc{\vij}{v_{ij}}
\nuc{\vbi}{\bar v_i}
\nuc{\vbj}{\bar v_j}
\nuc{\Vi}{V_i}
\nuc{\Vj}{V_j}
\nuc{\Vij}{V_{ij}}
\nuc{\Vbi}{\bar V_i}
\nuc{\Vbj}{\bar V_j}

\nuc{\wt}{w_t}
\nuc{\wtmw}{w_{t-1}}
\nuc{\wicj}{w_{i,j}}
\nuc{\wicjclcm}{w_{i,j,l,m}}

\nuc{\xwn}{x_1^n}
\nuc{\xt}{x_t}
\nuc{\xtpw}{x_{t+1}}
\nuc{\xtmw}{x_{t-1}}
\nuc{\xo}{x_0}
\nuc{\xw}{x_1}
\nuc{\xto}{x_2}
\nuc{\xT}{x_T}

\nuc{\xTt}{x^T_t}
\nuc{\xtt}{x^t_t}
\nuc{\xtmwt}{x^{t-1}_t}

\nuc{\xtil}{\tilde{x}}
\nuc{\xtilt}{\tilde{x}_t}
\nuc{\xtilu}{\tilde{x}_u}
\nuc{\xtiltgs}{\tilde{x}_{t|s}}
\nuc{\xtilugs}{\tilde{x}_{u|s}}
\nuc{\xtiltgt}{\tilde{x}_{t|t}}
\nuc{\xtiltgT}{\tilde{x}_{t|T}}
\nuc{\xtilugT}{\tilde{x}_{u|T}}
\nuc{\xtiltgtmw}{\tilde{x}_{t|t-1}}
\nuc{\xhat}{\hat{x}}
\nuc{\xhatt}{\hat{x}_t}
\nuc{\xhattgs}{\hat{x}_{t|s}}
\nuc{\xhattgtmw}{\hat{x}_{t|t-1}}
\nuc{\xhattt}{\hat{x}_t^t}
\nuc{\xhattgt}{\hat{x}_{t|t}}
\nuc{\xhattgT}{\hat{x}_{t|T}}
\nuc{\xhattmwgtmw}{\xhat_{t-1|t-1}}
\nuc{\xtgTf}{x_{t|T}^f}
\nuc{\xtgTb}{x_{t|T}^b}
\nuc{\xtmwgTf}{x_{t-1|T}^f}
\nuc{\xtmwgTb}{x_{t-1|T}^b}
\nuc{\ytgTf}{y_{t|T}^f}
\nuc{\ytgTb}{y_{t|T}^b}
\nuc{\ytmwgTf}{y_{t-1|T}^f}
\nuc{\ytmwgTb}{y_{t-1|T}^b}

\nuc{\yt}{y_t}
\nuc{\yr}{y_r}
\nuc{\ywn}{y_1^n}
\nuc{\ywT}{y_1^T}
\nuc{\yws}{y_1^s}

\nuc{\zwT}{z_1^{2T}}
\nuc{\clrr}[1]{{\color{black}{#1}}}

\nuc{\KT}{Karamata Tauberian}

\nuc{\bR}{\mathbb R}
\nuc{\bGwt}{\bar G_{12}}
\nuc{\bpwt}{\bar p_{12}}
\nuc{\bgwt}{\bar g_{12}}
\nuc{\bqwt}{\bar q_{12}}

\nuc{\Gwt}{G_{12}}
\nuc{\pwt}{p_{12}}
\nuc{\gwt}{g_{12}}
\nuc{\qwt}{q_{12}}
\nuc{\cwt}{c_{12}}

\nuc{\Ls}{L_*}

\renewcommand{\theequation}{\arabic{section}.\arabic{equation}}

\newtheorem{theorem}{Theorem}
\newtheorem{lemma}{Lemma}
\newtheorem{result}{Result}
\newtheorem{corollary}{Corollary}[theorem]
\newtheorem{proposition}{Proposition}[theorem]

\nuc{\clrb}[1]{{\color{blue}{#1}}}

\begin{document}
\title{Asymptotic Classification Error for \\ Heavy-Tailed Renewal Processes}
\author{Xinhui Rong and Victor Solo
\footnote{Authors are with School of Electrical Eng. $\&$ Telecommunications, 
UNSW, Sydney, Australia. }  
}    
  
\maketitle

\begin{abstract}                          
Despite the widespread occurrence of classification problems 
and the increasing collection of point process data 
across many disciplines, 
study of error probability for \pp\ classification only emerged very recently. 
Here, we consider classification of renewal processes. 
We obtain asymptotic expressions for the \Bhatt\ 
bound on misclassification error probabilities for 
heavy-tailed renewal processes. 
\end{abstract}

\section*{Copyright Statement}
This work has been submitted to the IEEE for possible publication. Copyright may be transferred without notice, after which this version may no longer be accessible.

\mysection{Introduction}
Point processes have wide applications in, e.g. neural coding \cite{Bial97} and stochastic finance \cite{Tankov}. Recent arising areas include social media \cite{Zhou13} and event triggered state estimation \cite{Shi}. 

\clrr{While time-series classification is well studied (e.g., \cite{Taniguchi}), 
work on \pp\ classification has emerged only recently. 
The Victor-Purpura \cite{Victor97}, ISI \cite{Kreu07} and SPIKE \cite{Kreu11} distances 
enabled clustering of neural spike trains modeled as \pp es.  
Mixture model approaches to spike-train clustering have also gained traction \cite{Wang23,Tao18}, including machine-learning methods \cite{Lin19}. 
Łukasik et al. \cite{Lukasik16} applied Hawkes process classification to social media data. 
For two-dimensional point fields, the likelihood-ratio 
and k-nearest-neighbor rules for Poisson fields are compared in \cite{Cholaquidis15}, 
and Mateu et al. \cite{Mateu14} classified plant communities modeled as Poisson fields.}

\clrr{Despite these emerging applications, the error analysis lags heavily behind.} 
Our previous work \cite{xr21} had the first theory about 
bounding the \pp\ classification error probability 
and was followed by \cite{Pawl23a,Pawl23b}. 
However, analytical bounds are only derived for Poisson processes. 
For \rp es, we had to resort 
to numerical Laplace inversion and simulation. 
While this is useful, asymptotic analysis would enable one to draw very
clear general conclusions and avoid massive simulations. 
The asymptotic error rate also helps determine how long the observation period 
is needed to achieve efficient classification. 
In \cite{Rong24s}, we developed asymptotic bounds for regular \rp es 
where the moment generating functions (MGFs) of inter-event times (IETs) exist.

In this paper, we obtain the asymptotic 
\Bhatt\ bound (BB)\clrr{, a classic upper bound of the error probability,} for a general class of heavy-tailed renewal processes, 
where the IET MGFs do not exist 
and the IET survivor functions are eventually regularly varying. 
We consider binary classification of heavy-tailed \rp es (not fields) 
\clrr{and reveal the completely different behavior of the bound 
from the regular case \cite{Rong24s} using Laplace transform (LT) analysis.}
It must be noted that classifying \rp es 
is totally different from classifying 
independent and identically distributed (iid) sources, 
and from the tail analysis of sums of iid positive RVs.  
See next section for more details. 

The rest of the paper is organized as follows. 
In Section \ref{pre}, we provide preliminaries of 
\pp es and classification. 
In Section \ref{BB}, we give a key theorem on the LT of the BB. 
In Section \ref{hvy}, we obtain the asymptotic \Bb\ for heavy-tailed \rp es. 
Section \ref{sim} contains both analytical and numerical 
analysis of Pareto distributed \rp es. 
Section \ref{con} gives conclusions. 
The Appendix contains proofs. 
\mysection{Classification of Renewal Processes}
\label{pre}
Here we recap some basic background on the \rp\ \cite{Dale03} and 
the likelihood ratio classification. %
\Pp es are characterized completely differently from continuous-valued RVs, 
so the classification differs from usual. %

\subsection{Renewal Processes}
A point process is a record of a random number $N_T$ of random event times 
$\TwN=[\mat{T_1&T_2&\dotsm&T_{N_T}}]^\top, 0<T_{r-1}<T_r<T$, 
observed in a fixed, continuous time interval $(0,T]$. 
Note that the number $N_T$ of observed events is also a RV. 
This makes \pp\ a hybrid process with both 
continuous- and discrete-valued RVs. 

A \rp\ is a point process with iid positive IETs 
$X_r = T_r-T_{r-1}>0$. 
Suppose the common density of the IETs is $p(x)$. 
Introduce 
the cumulative distribution function $F(x) = \intox p(u)du$, 
the survivor function $S(x) = 1-F(x)$. 
Then, the likelihood function of a renewal process trajectory 
$\{N_T=n, \TwN=\twn=[\mat{t_1&t_2&\dotsm&t_{n}}]^\top\}$ is given by
\eq{
\cL(\twn,n) &= 
\cas{
S(T),&n=0\\
\sqbra{\sprod{r=1}^np(t_r-t_{r-1})} S(T-t_n),&n>0},
}
where we define $t_0=0$. 
The hybrid likelihood sums, integrates to $1$ as follows
$
1 
	= S(T) + \ssum{n=1}^\infty\int_{R_n(T)}\cL(\twn,n)d\twn,
$
where the region $R_n(T) = \{\twn| 0<t_1<t_2<\dotsm<t_n<T\}$. 
Later, we informally write the sum/integral as $\ints f(\twn,n)d\twn$ for a function $f$.
The above distinguishes \rp es from sums of iid positive RVs.

We introduce 
the hazard function $h(x) = \frac{p(x)}{S(x)}$, 
the integrated hazard $H(x) = \intox h(u)du$ 
and the hazard relations
\eqn{
\label{hr}
p(x) = h(x)e^{-H(x)},\quad S(x) = e^{-H(x)}.
}
Using these relations, 
one can also derive the same \rp\ likelihood function 
from the Janossy density \cite{Dale03} of the general \pp es (see e.g. \cite{xr21}).

\subsection{Likelihood Classification and the Misclassification Error}
Here we consider binary statistical classification of a \rp, i.e. 
given a recording $\{N_T=n, T_1^{N_T}=\twn\}$ of a \rp\ over a fixed time interval $(0,T]$, 
we classify it into one of the $M=2$ classes. 
The class label $C\in\{1,2\}$ is a RV with mass function $\Pr[C=k]=\pi_k$ and $\pi_1+\pi_2=1$. 
We denote the class IET density $p_k(x)$ and 
the corresponding class \rp\ likelihood $\cL_k(\twn,n), k=1,2$.

A classifier $\hat C$ takes value $k$ to 
assign the class label $k$ to a given trajectory $\{n,\twn\}$. 
The misclassification error probability is defined as
$
e(T) 	= \Pr[\hat C\neq C] = \ssum{k}\pi_k\Pr[\hat C\neq k|C=k]. 
$
Assuming that the priors $\pi_1,\pi_2$ and 
the likelihoods $\cL_1, \cL_2$ are known, 
the likelihood classifier (a.k.a. Bayes classifier) minimizes \cite{DeVr97}
the misclassification error probability by \clrr{assigning 
\eq{
\mbox{$\hat C=k$ if }\pi_k\cL_k(\twn,n) > \pi_j \cL_j(\twn,n),\quad k,j\in\{1,2\}, 
}
and the misclassification error probability is minimized as
\eq{
P_e(T) 
=& \pi_1\Pr[\hat C=2|C=1] + \pi_2\Pr[\hat C=1|C=2]\\
=& \pi_1 \Pr[\pi_2\cL_2>\pi_1\cL_1|C=1] \\
&\quad+ \pi_2\Pr[\pi_2\cL_2<\pi_1\cL_1|C=2]\\
=& \ints\min\{\pi_1\cL_1,\pi_2\cL_2\}d\twn. 
}
We note the error probability $P_e(T)$ is a function of $T$.}

\section{{The Point Process \Bhatt\ Bound}}
\label{BB}
The classic BB is an upper bound on 
the error probability 
as a direct result of the inequality $\min\{a,b\} \leq \sqrt{ab}$ \cite[Section 3.4]{DeVr97}. 
For \pp es, the BB $P_b(T)$ takes the form
\eq{
P_b(T) &= \sqrt{\pi_1\pi_2}B(T),\quad
B(T) = \ints\sqrt{\cL_1\cL_2}d\twn.
}

\clrr{
We choose the BB for the following reasons. 
\blist{
\item Exact calculation of the error probability is 
only possible for some Poisson processes \cite{xr21}. 
\item The \Bb\ is shown to bound $P_e(T)$ reasonably well \cite{xr21}. 
\item The Chernoff bound is possibly tighter, 
replacing the integrand in $B(T)$ with $\cL_1^\lambda \cL_2^{1-\lambda}$ with the optimal $\lambda$. 
However, its analysis follows analogously from that of the BB. 
The Shannon bound is also related to $B(T)$ \cite{Rong24s}. 
\item The lower error bound also relates to the BB. 
$P_e(T)$ is lower bounded by $\wo2(1-\sqrt{1-4B^2(T)})$ 
\cite[Section 3.4]{DeVr97}\footnote{
We thank a reviewer for pointing this out.}. Using the identity 
$1-\sqrt{1-x} = 
\frac{x}{1+\sqrt{1-x}}$, we find the lower bound to decay at an order $B^2(T)$.
}}

Finding the analytic formula for the BB for \rp es is hard 
given the unusual likelihood. 
However, in \cite{xr21}, we evaluated the \Bb\ in terms 
of repeated convolution and 
found the analytic LT of the \Bb\
as given below. 

\theo{\cite{xr21}
For \rp es, $B(T)$ has LT
\eq{
\bar B(s) = \frac{\bar G_{12}(s)}{1-\bar p_{12}(s)},
}
where $\bpwt(s)$ is the LT of $\pwt(x) = \sqrt{p_1(x)p_2(x)}$ and 
$\bGwt(s)$ is the LT of $\Gwt(x) = \sqrt{S_1(x)S_2(x)}$, 
and $S_k(x)$ is the survivor function of class $k$ IETs.
}

\clrr{The explicit LT representation enables asymptotic analysis of the error bound.}
It is straightforward to check that  
the LTs $\bGwt(s)$ and $\bpwt(s)$ always 
exist for $s\in(0,\infty)$. 
Analysis on the positive half-line is sufficient for asymptotic analysis. 
The distributions where $\bpwt(s)$ also exists for some $s<0$ 
are called regular and are dealt with in \cite{Rong24s} in a multiclass setting.

Note that $\Gwt(x)=\sqrt{S_1(x)S_2(x)}$ is also a survivor function. 
For future use, we introduce the corresponding density $\qwt(x)$. 
Noting the hazard relations (\ref{hr}), we have 
\eqn{\label{q12}
{\qwt(x) = \ddx(-\Gwt(x))	= \frac12(h_1(x)+h_2(x))\Gwt(x).}
}

\mysection{Asymptotic Bhattacharyya Bound}
\label{hvy}
We focus on a subclass of heavy-tailed IET distributions, 
where the survivor functions are eventually regularly varying,
\clrr{and derive the asymptotic BB.}
The class of such regularly varying distributions 
is also a subset of the sub-exponential distributions \cite{Foss11}.
We first define the regularly varying assumption 
and then present our main theorem.

{\bf Definition 1. }{\it Asymptotic relations. }
{A function $f$} is asymptotically equivalent to $g$ at $\tau$, 
or $f(\tau)\sim g(\tau)$, with $\tau$ finite or infinite, if 
$
\lim_{t\to\tau} \frac{f(t)}{g(t)}=1.
$

{\bf Definition 2. }{\it Slowly varying functions.} \cite{Fell88}
{A function $L$, defined on the positive half-line, is slowly varying (at $\infty$) 
if for every $t$, we have $\frac{L(xt)}{L(x)}\to1$ as $x\toi$. }

{\bf Assumption A1. }{\it Regularly varying survivors.} 
For each class $k$, the survivor function 
$S_k(x) \sim x^{-\rho_k}L_k(x),\rho_k>0$, as $x\toi$,
where $L_k$ is slowly varying.

We will use the following properties and the \KT\ (KT) theorem
{relating asymptotics in the Laplace domain to the time domain.}

{\it Lemma 1. }
\cite[Proposition 1.3.6]{Bing87}
Suppose $L_1(x)$ and $L_2(x)$ are slowly varying. Then, 
\nlist{
\itum{a} $\limxi x^{-\beta}L_1(x) = 0, \beta>0$, 
\itum{b} $L_1^\beta(x),\beta\in\bR$ is slowly varying, and 
\itum{c} $L_1(x)L_2(x)$ and $L_1(x)/L_2(x)$ are slowly varying.
}

\theo{
\label{TA}\cite[Section XIII.5]{Fell88} 
\KT\ (KT) Theorem. 
Let $\bar S(s)$ be the LT of a survivor function $S(x)$, 
$U(x)=\intox S(u)du$ and $\Gam(x)$ be the Gamma function. 
Consider the asymptotic relations
\eq{
\arr{lrcll}{
\mbox{(a) } & S(x) &\sim& \frac1{\Gamma(1-\rho)}x^{-\rho}L(x), &\mbox{as } \xtoi\\
\mbox{(b) } & U(x) &\sim& \frac1{\Gamma(2-\rho)}x^{1-\rho}L(x), &\mbox{as } \xtoi\\
\mbox{(c) } & \bar S(s) &\sim& \wo{s^{1-\rho}}L(\frac1s), &\mbox{as } \stoo.
}}
We have (a) $\LRa$ (b) for $\rho<1$, 
(a) $\LRa$ (c) for $\rho<1$, and
(b) $\LRa$ (c) for $\rho\leq1$.
}

The KT theorem is sophisticated and 
has been generalized over a century \cite{Fell88}\cite{Bing87}. 
To keep the discussion and conditions simple, we use 
the most basic KT theorem above. The more general
KT theorem requires additional technical features that
are not needed here. We present the main theorem below.

\theo{
\label{Bhvy}
Under assumption A1, as $T\toi$, 
\eqn{
\label{Ba}
B(T)\sim B_*(T)=\frac1{1-c_{12}}T^{-\rho}L(T), 
}
where $\rho = \frac{\rho_1+\rho_2}2$,  
$c_{12} = \bpwt(0) = \intoi\pwt(x)dx$, and 
$L(T)=\sqrt{L_1(T)L_2(T)}$ %
is slowly varying.
}

The proof is given in Appendix. 
The BB has near power decay, 
slower than the exponential decay in the regular IET case \cite{Rong24s}, 
which means that heavy-tailed \rp es are harder to classify. 
\clrr{Further, the decay exponent $\rho$ of BB is the average of the tail indices $\rhok$ 
of the IET survivor functions.}

The assumption of known priors and likelihood functions is somewhat restrictive. 
In practice, one would `plug in' estimates of model parameters from a training data set. 
Then the error bound would depend on the asymptotic properties 
of these estimators. 
It is possible that under some technical conditions, the new bound and 
the one given above will be asymptotically equivalent. 
However, studying that is a challenging problem and will be pursued elsewhere.

\mysection{Analytical and Numerical Studies}
\label{sim}
\clrr{For insight into the asymptotic misclassification error, 
we treat Pareto IETs as an example and 
derive a specialized form of Theorem \ref{Bhvy}. 
We offer both analytical and numerical studies.}

\subsection{Pareto Example}
We assume the class IETs obey Pareto distribution supported on $[\alpk,\infty)$ 
and the class density is given by $p_k(x) = \frac{\alpk^{\betk}\betk}{x^{\betk+1}}, x\geq\alpk$, 
with $\alpha_2\geq\alpha_1>0, \betk>0$. 

To interpret  
the asymptotic BB, we introduce $\beta = \frac12(\beta_1+\beta_2), \alpha = \alpha_1^{\beta_1/2}\alpha_2^{\beta_2/2}$ and two unit-free
independent parameters
$
\gam =\beta_1/\beta_2$ and $\theta = \sqrt{\alpha_1/\alpha_2}\leq1.
$
Then, applying Theorem \ref{Bhvy}, the asymptotic expression is given by 
\eqn{
\label{Bsuf}
B_*(T)=\frac1{(T/\alpha_2)^\beta}\frac{\theta^{\beta_1}}{1-\delta\theta^{\beta_1}},\quad \delta = \frac{2\sqrt\gam}{1+\gam}\leq1.
}
Differentiating w.r.t. $\theta$ shows that $B_*(T)$ is an increasing function of $\theta$. 
Clearly, then the closer $\theta$ is to 1 
the more conservative the asymptotic expression is
and the harder the classification is. 
This also applies to the case when $\gam$ approaches $1$. 
Further, the closer $\beta_1$ is to 0, i.e. the heavier the tail, 
the harder the classification is. 
Also note that larger $\alpha_2$ means that 
we need longer recording time to get a better classification. 
$B_*(T)$ is also an increasing function of $\delta$.

\subsection{Pareto Simulations}
We simulate \rp es with Pareto distributed IETs to compare the asymptotic \Bb\ $\sqrt{\pi_1\pi_2}B_*(T)$ in (\ref{Bsuf}) with
\nlist{
\item Monte-Carlo \Bb\ $\sqrt{\pi_1\pi_2}\hat B(T)$, and 
\item Monte-Carlo error probability $\hat P_e(T)$. 
}

We fix $\alpha_1 = 10, \beta_1 = 1$ and the priors $\pi_1 = \pi_2 = 0.5$. 
We vary $T,\alpha_2,\beta_2$, or equivalently 
$T,\theta,\gamma$ taking the unit-free expression. 
We vary  $[T,\theta,\gamma]$ 
in a $15\times3\times5$ log-scaled range of 
$[2\times10^2\sim2\times10^4, 2^{-0.5}\sim2^{-0.1}, 2^{-1}\sim2]$.

\subsubsection{Comparison with the bound} 
In \cite{xr21}, we showed the numerical inversions of LTs 
behave badly for large $T$. 
Therefore, we sample the Monte-Carlo \Bb\ $\hat B(T)$ 
as follows. For each parameter grid, 
we simulate $M=10^7$ realizations of \rp es with \rp\ likelihood 
$\cL_q(\twn,n) = \sprod{r=1}^n\qwt(x_r)\Gwt(T-t_n)$ 
with IET density $\qwt$ {defined in (\ref{q12})}. 
Let $n_m$ be the number of events for $m$-th realization. 
We sum $\delta^{n_m}$, 
only for realizations where all IETs $>\alpha_2$ or {when }$n_m=0$.  
Then divide the summation by $M$ to get $\hat B(T)$. 
Limit of space precludes derivation of the above sampling method. 

We simulate the \rp es conveniently by 
repeatedly generating IETs by inverse transform sampling 
until the next event time exceeds $T$. 
In simulations, we observe that for the largest $T=2\times10^4$, 
the smallest $\hat B(T)$ is $6.7\times10^{-5}$, happening when $\gam=0.5, \theta=0.71$, 
while the largest $\hat B(T)$ is $0.031$ happening when $\gam=2,\theta=0.93=2^{-0.1}$. 

We plot the $15\times3\times5$ sets of 
$y=\ln[\hat B(T) / B_*(T)]$ against $\gamma,\theta$ and $\tau=\ln(T/\alpha_2)$ 
in the heat map in Fig. 1. 
We expect $y(\tau) \to0$ as $\tau\to\infty$. 
From Fig. 1, we have the following observations. 
(a) $y(\tau)\to0$ for large $\tau$ as expected.  
(b) The \Bb\ $B(T)$ reaches its asymptotics $B_*(T)$ faster 
for small $\theta$ and for $\gamma$ away from $1$, 
i.e. when class $1$ and $2$ are easier to distinguish.  
(c) We have to sample $10^7$ realizations to estimate $B(T)$ 
reasonably well. However, the asymptotic expression $B_*(T)$ 
is straightforward and much easier to interpret.

\subsubsection{Comparison with the error probability} 
We have demonstrated our theorem by comparing $B_*(T)$ with $\hat B(T)$. 
However, one would also be interested in  
whether the error probability $P_e(T)$ also has power decay like the bound. 
Hence, we simulate $M=2\times10^7$ realizations of \rp es 
for each class and 
carry out likelihood classification 
to sample the Monte-Carlo error probability $\hat P_e(T)$.

In simulations, for the largest $T=2\times10^4$, 
the smallest error probability $\hat P_e=3.9\times10^{-6}$ 
happening when $\gam=0.5,\theta=0.81$,  
while the largest error probability $\hat P_e=2.1\times10^{-3}$ 
happening when $\gam=1.4,\theta=0.93=2^{-0.1}$.

We plot the heatmap of $z=\frac{\ln[2\hat P_e(T)/B_*(T)]}{\beta\ln(T)}$ 
against $\gamma,\theta$ and $\tau=\ln(T/\alpha_2)$ in Fig. 2. 
$z$ is an asymptotic relative discrepancy 
between the true error decay profile 
and the \Bb\ decay profile. 
If $P_e(T)$ also has power decay, 
we expect $z\to$ constant $\leq0$ as $\tau\toi$; 
Fig. 2 confirms such property. 
We also have the following observations. 
(a) For small $\tau$ (or $T$), some $z\geq0$ because the asymptotic \Bb\ $B_*(T)$ 
does not necessarily bound the error probability for small $T$. 
However, the empirical \Bb\ $\hat B(T)$ does (see \cite{xr21}). 
(b) For $\gam\geq1$, $z$ has smaller absolute values for larger $\theta$. 
This indicates that $B_*(T)$ bounds the error decay profile better 
when $\alpha_1$ and $\alpha_2$ are closer. 
But for $\gam<1$, the conclusion is the converse. 
(c)  For fixed $\theta$, $B_*(T)$ bounds the error decay profile best for $\gam\approx1$. 
(d)  For $\gam=0.5$, $z$ has not converged 
even with our largest observation time $T=2\times10^4$; 
we need even longer observation time and thus also more realizations 
to sample the error probability decay profile. 
However, our asymptotic \Bb\ result is simple and takes almost no time to calculate.

\mysection{Conclusions}
\label{con}
In this paper, we developed, for the first time, 
the asymptotic approximation $B_*(T)$, of the \Bhatt\ bound (BB) 
for classifying \rp es 
with heavy-tailed inter-event time distributions, 
based on 
Laplace transform analysis. 
We found that the \Bb\ has near power decay,  
indicating that the heavy-tailed \rp es are harder to classify 
than the regular \rp es whose \Bb\ decays exponentially. 
In the simulation studies, 
we compared $B_*(T)$ with both 
the empirical \Bb\ $\hat B(T)$ and the empirical error probability $\hat P_e(T)$. 

In the future, we will consider the `plugged-in' likelihoods, 
the general subexponential distributions, 
Hill-horror distributions and multiclass classification.

\begin{figure}[t]
\hfill
\begin{minipage}[t]{1\linewidth}
  \centering
  \centerline{\includegraphics[width=12cm]{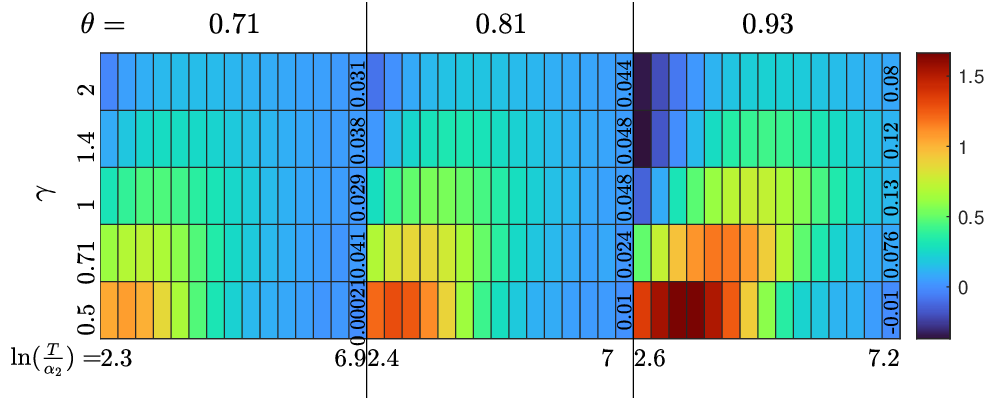}}
\end{minipage}
\caption{Heatmap of $y=\ln[\hat B(T)/B_*(T)]$: %
We convert $15\times3\times5$ sets of results in one heatmap. 
We display the values of $y$ 
for the largest $T$ 
for each pair of $(\gamma,\theta)$. 
It is verified that $y\to0\Ra B(T)\sim B_*(T)$. 
}
\end{figure}

\begin{figure}[t]
\hfill
\begin{minipage}[t]{1\linewidth}
  \centering
  \centerline{\includegraphics[width=12cm]{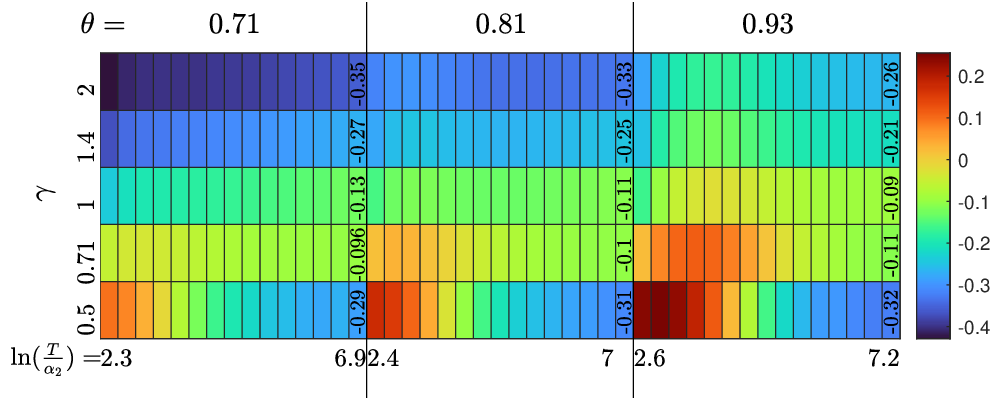}}
\end{minipage}
\caption{Heatmap of $z=\frac{\ln[2\hat P_e(T)/B_*(T)]}{\beta\ln(T)}$: 
We demonstrate that the classification error probability $P_e(T)$ also has power decay, 
by showing that $z$ converges to a negative constant. 
}
\end{figure}

\section*{Appendix. Proof of the Main Theorem}
\renewcommand{\theequation}{A.\arabic{equation}}
Here, we prove Theorem \ref{Bhvy}. 
Under A1 and by Lemma 1, we have
\eqn{
\label{Ga}
\Gwt(x) \sim x^{\rho-1}L(x).
}
{We prove Theorem 3 in three steps, based on the value of $\rho$. }

{\bf Step 1: $0<\rho<1$}
The asymptotic relation (\ref{Ba}) is 
a direct application of the KT theorem. 

{\it Result 1:} 
The relation (\ref{Ba}) holds when $0<\rho<1$.

{\it Proof. }
For $0<\rho<1$, the relation (\ref{Ga}) and Theorem \ref{TA} give
$
\Gwt(x)\sim x^{\rho-1}L(x)\Ra \bGwt(s) \sim \frac{\Gam(\rho)}{s^\rho}L(\frac1s).
$ 
Then, note that $\bpwt(0) = \intoi \pwt(x)dx = \cwt$. We have 
$
\bar B(s) = \frac{\bGwt(s)}{1-\bpwt(s)} \sim \frac{\Gam(\rho)}{(1-\cwt)s^\rho}L(\frac1s),0<\rho<1, 
$
as $\stoo$. 
Now use Theorem \ref{TA} again to get (\ref{Ba}). 
\hfill$\square$

{\bf Step 2: $\rho=0$}
This case is not straightforward 
since (a) and (b) in Theorem \ref{TA} do not imply each other when $\rho=0$. 
Thus, instead, we analyze the LT of $P(T)=TB(T)$. 
The LT exists on $s\in(0,\infty)$ since $P(T)$ is bounded by a polynomial with finite power. 
We need the following lemmas.

{\it Lemma A: } Let $L(x)$ be slowly varying and locally bounded on $[a,\infty),a>0$ 
and $S(x)\sim x^{-1}L(x)$ at $\infty$. Then, 
$U(x) = \intox S(u)du \sim L_*(x)$ at $\infty$, 
where $L_*(x)=\int_a^x u^{-1}L(u)du$ is slowly varying. 

{\it Proof.} 
This is a generalization of \cite[Proposition 1.5.9a]{Bing87}. 
The proof is simple and thus omitted.
\hfill$\square$

{\it Lemma B: } $\pwt(x)\leq\qwt(x)${, 
where $\qwt$ is the density function whose survivor function is $\Gwt$ and is given in (\ref{q12})}. 

{\it Proof.}
Use the hazard relations (\ref{hr}) 
to find 
$\pwt(x)=\sqrt{h_1(x)h_2(x)}\Gwt(x)\leq \frac{h_1(x)+h_2(x)}{2}\Gwt(x) = \qwt(x)$.\hfill$\square$

{\it Result 2: }
The relation (\ref{Ba}) holds when $\rho=0$.

{\it Proof.}
Introduce $P(T) = TB(T)$. 
Then its LT
$
\bar P(s) = \intoi e^{-sT}TB(T)dT = -\bar B'(s) 
	= \frac{-\bGwt'(s)}{1-\bpwt(s)} + \frac{\bpwt'(s)\bGwt(s)}{(1-\bpwt(s))^2}, s>0
$. 
We have $\bpwt(0)=\cwt$. So we examine $\bGwt'(s),\bpwt'(s)$ and $\bGwt(s)$. 
First, $-\bGwt'(s)$ is the LT of $x\Gwt(x)$ and 
$x\Gwt(x)\sim L(x)$, as $\xtoi\Ra -\bGwt'(s)\sim \frac1sL(\frac1s)$, as $\stoo$. 
Now use Lemma A and Theorem \ref{TA} to get
$U(x) = \intox \Gwt(u)du \sim L_*(x){=\intox u^{-1}L(u)du}$, as $\xtoi\Ra\bGwt(s)\sim \bar L_*(\frac1s)$, as $\stoo$, 
where $\Ls(x)$ is slowly varying due to Lemma A{ and $\bar L_\ast(s)$ is the LT of $\Ls(x)$}. 
Then, due to Lemma B, we have for real $s>0$, 
$-\bpwt'(s) = \intoi e^{-sx}x\pwt(x)dx
	\leq \intoi e^{-sx}x\qwt(x)dx 
	= \intoi e^{-sx}xd(-\Gwt(x))
	= 0 + \intoi\Gwt(x)(e^{-sx}-sxe^{-sx})dx
	= \bGwt(s) + s\bGwt'(s)$.  
This indicates that for $s>0$, 
\eq{
\frac{-\bpwt'(s)\bGwt(s)}{\frac1sL\bra{1/s}}&\leq\frac{[\bGwt(s)+s\bGwt'(s)]\bGwt(s)}{\frac1sL\bra{1/s}}\sim {s\hat L(1/s),}
}
{where $\hat L(\wos)=\frac{L_*\bra{1/s}\sqbra{L_*\bra{1/s}-L\bra{1/s}}}{L\bra{1/s}}$ 
is slowly varying and so we conclude $s\hat L(1/s)\to0$ by Lemma 1.}

We thus find that $-\bGwt'(s)$ dominates $\bar P(s)$, so that 
$\bar P(s)\sim\frac{1}{1-\cwt}\frac1sL(\frac1s)$, as $\stoo\Ra P(T)\sim\frac{1}{1-\cwt}L(T)$, as $\Ttoi\Ra B(T) \sim\frac{1}{1-\cwt}T^{-1}L(T)$, as $\Ttoi$,
as needed.
\hfill$\square$

{\bf Step 3: $\rho<0$} 
The KT theorem does not cover this case. 
We now need to analyze the LT of $P_m(T) = T^mB(T)$,  
where the integer $m$ satisfies
\eq{
x^m\Gwt(x) \sim x^{\beta-1}L(x),\quad 0\leq\beta=\rho+m<1.
}
The LT also exists by the same arguments above. 
In this case, we need more results about the non-dominating LT terms. 
Let $f^{(n)}$ be the $n$-th derivative of $f$. 
The following lemmas hold. 

{\it Lemma C: } 
For all integers $n<m$, $(-1)^n\bGwt^{(n)}(0)=\intoi x^n\Gwt(x)dx<\infty$. (cf. \cite[Proposition 1.5.10]{Bing87})

{\it Lemma D: } 
For all integers $n\leq m$, $(-1)^n\bpwt^{(n)}(0)=\intoi x^n\pwt(x)dx<\infty$.

{\it Proof.}
{It suffices to consider
$\intoi x^m\pwt(x)dx\leq\intoi x^m\qwt(x)dx
= \intoi x^md(-\Gwt(x))
= -x^m\Gwt(x)\vert_0^\infty + m\intoi\Gwt(x)x^{m-1}dx
= m(-1)^{m-1}\bGwt^{(m-1)}(0)<\infty$.
The inequalities follow from Lemma B and Lemma C.} 
\hfill$\square$

{\it Result 3: }
The relation (\ref{Ba}) holds when $\rho<0$.

{\it Proof.}
We need to discuss two cases. 

First, for $0<\beta<1$, 
let $P_m(T) = T^mB(T)$. 
Its LT has the term
$\frac{(-1)^m\bGwt^{(m)}(s)}{1-\bpwt(s)}$, 
and all other terms have finite limits as $\stoo$ due to Lemma C and Lemma D. 
Then following the same as in the proof of Result 1 gives the quoted result. 

Now consider $\beta=0$ and $P_{m+1}(T) = T^{m+1}B(T)$. 
We need to examine its LT terms with 
$\bGwt^{(m)}, \bGwt^{(m+1)}$ and $\bpwt^{(m+1)}$, 
which are possibly infinite as $\stoo$. 
{Following parallel steps from the proof of Result 2}, 
using integral by parts to get, for $s>0$, 
$(-1)^n\bpwt^{(n)}(s) \leq \bGwt(s) + (-1)^{n-1}s\bGwt^{(n)}(s)$. 
We can also get that $\bGwt^{(m)}(s)$ is slowly varying but 
$(-1)^{m+1}\bGwt^{(m+1)}(s)\sim \frac1sL(\frac1s)$. 
Then, we have that $\bpwt^{(m+1)}(s)$ is slowly varying. 
However, the only term containing $\bGwt^{(m+1)}$ is 
$\frac{(-1)^{m+1}\bGwt^{(m+1)}(s)}{1-\bpwt(s)}$, 
so it still dominates. The result then also follows easily.
\hfill$\square$

Now Results 1-3 finally prove Theorem \ref{Bhvy}.


\bibliographystyle{IEEEtran}
\bibliography{bib/xr-bib}  
\end{document}